\pdfoutput=1
\documentclass[10pt,twocolumn,letterpaper]{article}

\usepackage[pagenumbers]{iccv}
\usepackage{marvosym} 

\usepackage{colortbl}

\definecolor{iccvblue}{rgb}{0.21,0.49,0.74}
\usepackage[pagebackref,breaklinks,colorlinks,allcolors=iccvblue]{hyperref}
\usepackage{multirow}
\usepackage[T1]{fontenc}
\usepackage{times}
\usepackage{colortbl}
\definecolor{lightblue}{HTML}{e6f3ff}
\definecolor{lightpink}{HTML}{fff0f0}

\newcommand{\ModelName}{LangBridge\xspace}

\usepackage{colortbl}

\def\paperID{13723} 
\def\confName{ICCV}
\def\confYear{2025}

\title{\ModelName: Interpreting Image as a Combination of Language Embeddings}

\author{Jiaqi Liao$^{1*}$, Yuwei Niu$^{6,8*}$, Fanqing Meng$^{5,1*}$, Hao Li$^{2,1}$, Changyao Tian$^{2,1}$, Yinuo Du \\Yuwen Xiong$^{1}$, Dianqi Li,
Xizhou Zhu$^{3,4}$, Li Yuan$^{6,7}$, Jifeng Dai$^{3,1\textsuperscript{\Letter}}$, Yu Cheng$^{2\textsuperscript{\Letter}}$ \\
{\normalsize\centering$^{1}$ Shanghai AI Laboratory}
{\normalsize\centering$^{2}$ The Chinese University of Hong Kong}
{\normalsize\centering$^{3}$ Tsinghua University}
{\normalsize\centering$^{4}$ SenseTime Research}
\\
{\normalsize\centering$^{5}$ Shanghai Jiao Tong University}
{\normalsize\centering$^{6}$ Peking University}
{\normalsize\centering$^{7}$ PengCheng Laboratory}
{\normalsize\centering$^{8}$ Chongqing University}
}

\begin{document}
\maketitle  
{\let\thefootnote\relax\footnote{$^*$Equal contribution. 
\textsuperscript{\Letter} 
}}
\begin{abstract}

\vspace{-11pt}
Recent years have witnessed remarkable advances in Large Vision-Language Models (LVLMs), which have achieved human-level performance across various complex vision-language tasks. Following LLaVA's paradigm, mainstream LVLMs typically employ a shallow MLP for visual-language alignment through a two-stage training process: pretraining for cross-modal alignment followed by instruction tuning. While this approach has proven effective, the underlying mechanisms of how MLPs bridge the modality gap remain poorly understood. Although some research has explored how LLMs process transformed visual tokens, few studies have investigated the fundamental alignment mechanism. Furthermore, the MLP adapter requires retraining whenever switching LLM backbones. To address these limitations, we first investigate the working principles of MLP adapters and discover that they learn to project visual embeddings into subspaces spanned by corresponding text embeddings progressively. Based on this insight, we propose LangBridge, a novel adapter that explicitly maps visual tokens to linear combinations of LLM vocabulary embeddings. This innovative design enables pretraining-free adapter transfer across different LLMs while maintaining performance. Our experimental results demonstrate that a LangBridge adapter pre-trained on Qwen2-0.5B can be directly applied to larger models such as LLaMA3-8B or Qwen2.5-14B while maintaining competitive performance. Overall, LangBridge enables interpretable vision-language alignment by grounding visual representations in LLM vocab embedding, while its plug-and-play design ensures efficient reuse across multiple LLMs with nearly no performance degradation. See our project page at \url{https://curryx-001.github.io/LangBridge.github.io/}.

\vspace{-12pt}

\end{abstract}      
\section{Introduction}
\label{sec:intro}
\begin{figure*}[h]
    \centering
    \includegraphics[width=\textwidth]{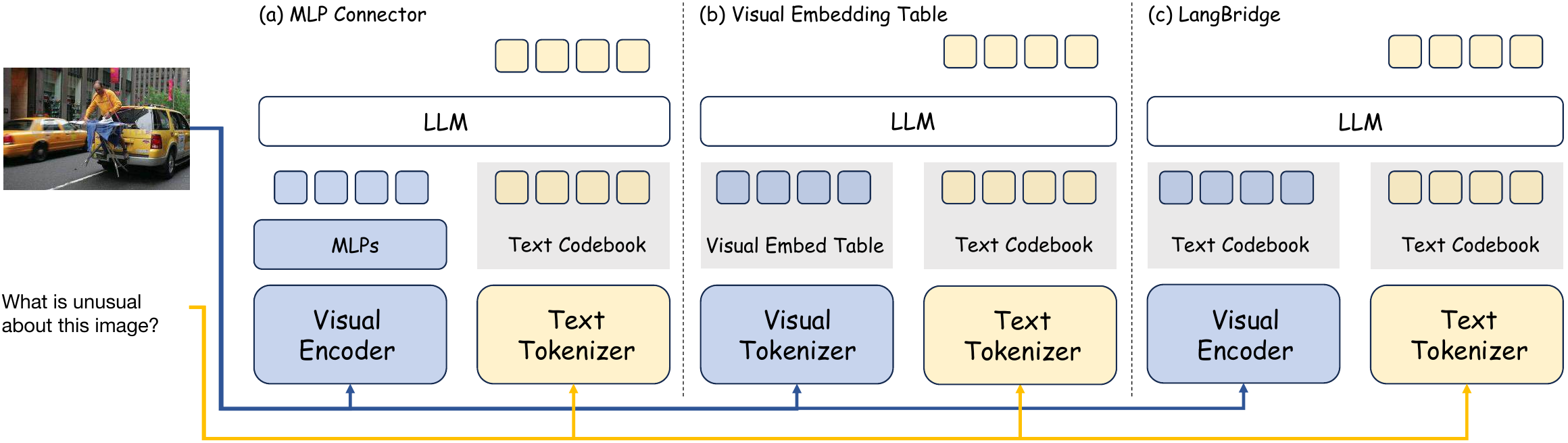}
    \vspace{-0.3mm}
    \caption{\textbf{Comparison of different connector types in LVLM: }(a) The MLP directly maps visual features into the LLM’s text embedding space. (b) The Ovis method uses a visual embedding table to produce structural visual embeddings and align the modalities. (c) \ModelName decomposes visual features into weighted combinations of LLM’s vocabulary vectors to form the visual embeddings}
    \vspace{-0.3mm}
    \label{fig:comparision}
\end{figure*}

Recent years have witnessed remarkable progress in Large Vision-Language Models (LVLMs)~\cite{alayrac2022flamingo, liu2024improved, zhu2023minigpt,lu2024deepseek,li2024mini,chen2024far,yang2024qwen2,lin2024moe}, which have demonstrated impressive capabilities in complex vision-language tasks, showcasing sophisticated understanding and reasoning across both visual and textual modalities. As a result, these models are increasingly being deployed in real-world applications, ranging from autonomous driving to robotics~\cite{sima2023drivelm, brohan2023rt, narayanaswamy2024using}.

As shown in Figure~\ref{fig:comparision} (a), mainstream Large Vision-Language Models follow the architecture of LLaVA~\cite{liu2024improved}, leveraging Vision Transformers (ViTs) for visual feature extraction and Multi-Layer Perceptrons (MLPs) to transform these features into visual embeddings aligned with the LLM’s text embedding dimension. These models also adopt LLaVA’s two-stage training process: (1) MLP warm-up for cross-modal alignment and (2) full-model instruction tuning. While this approach has proven effective, two critical challenges remain: 1) the underlying mechanisms of how MLPs bridge the modality gap are poorly understood. Although some research has explored how LLMs process transformed visual tokens~\cite{kaduri2024s,zhang2024large}, few studies have investigated the fundamental alignment mechanisms of MLPs, and 2) the MLP adapter requires retraining whenever switching LLM backbones due to input dimension mismatch and distributional shifts in feature representations.

To address these limitations, we systematically investigate the working principles of MLP adapters through two critical aspects: 1) how visual semantics are encoded in transformed embeddings, and 2) how MLPs learn cross-modal alignment during training. Our methodology employs visual analysis across four distinct training stages: 1) Pretrain-100 steps, 2) Pretrain-1000 steps, 3) Pretrain-2000 steps, and 4) SFT-tuned Model. For each stage, we extract visual features using Vision Transformers, transform them through stage-specific MLPs into visual embeddings, and compute cosine similarities between these visual embeddings and text embeddings. Through this process, we identify the top semantically related text tokens for each visual embedding at each training phase.

\begin{figure*}[h]
    \centering
    \includegraphics[width=\textwidth]{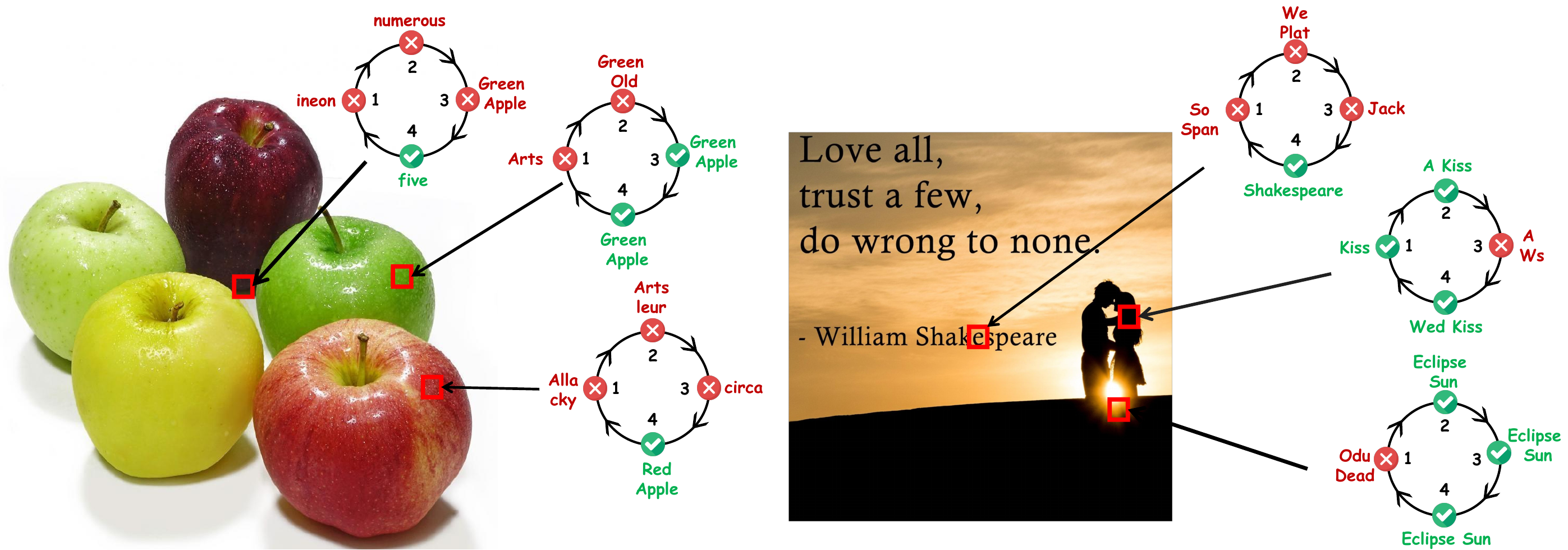}
    \vspace{-0.3mm}
    \caption{\textbf{Progressive semantic alignment in MLP adapters across training stages.} Circular graphs demonstrate the evolution of visual-text token associations through four training phases (Pretrain-100, Pretrain-1000, Pretrain-2000, and Final-SFT). (a) For a bunch of apples image, the MLP progressively refines associations from the meaningless text (``numerous", ``Arts") to meaningful text (``five", ``Green Apple"). (b) For a sunset silhouette scene and poem,  semantic mappings evolve from the meaningless text (``We Plat", ``Kiss") to contextually relevant tokens (``Shakespeare", ``Wed Kiss"), illustrating the MLP's increasing capability to project visual features into LLM's text embedding space.}
    \vspace{-0.3mm}
    \label{fig:Explainable}
\end{figure*}

As visualized in Figure~\ref{fig:Explainable} through circular graphs where nodes represent different training phases spanning from early to late stages (e.g. 1- 4), we can draw two key conclusions: 1. Visual embeddings demonstrate strong correlations with semantically-related text tokens. As shown in Figure~\ref{fig:Explainable}, green apple image patch exhibits high similarity with text tokens such as ``Green Apple". Similarly, the silhouetted sunset scene correlates with concepts like ``Eclipse Sun". This suggests that MLPs effectively map visual features into the text embedding space of LLMs, particularly in regions close to the semantic spaces formed by the corresponding text embeddings. 2. The projection capability of MLPs develops progressively during training. Figure~\ref{fig:Explainable} illustrates this evolution through multiple semantic pathways: from ``numerous" to "five", from "Arts" to ``Green Apple", and in the right scene, from ``We Plat" to ``Shakespeare", from ``Kiss" to ``Wed Kiss". These relationships become increasingly well-defined through the training stages, confirming our hypothesis that MLPs gradually learn to map visual features into regions proximate to their corresponding text embedding subspaces.

Building upon this insight, we introduce an explicit transformation approach called \textbf{Language Basis Vector Projection}. This approach explicitly projects visual features into the text embedding subspace of the LLM by representing them as linear combinations of the LLM's vocabulary embeddings. Based on this theoretical foundation, we propose \textbf{\ModelName}, a novel adapter architecture. The adapter first maps visual features to probability distributions that capture their semantic similarity with vocabulary text embeddings in the LLM. Subsequently, it generates the final visual embeddings by combining these text embeddings, weighted according to the computed probabilities. This innovative design elegantly addresses the challenge of model retraining when switching between different LLM backbones. The key insight is that our adapter only learns the linear combination relationship between visual patches and vocabulary embeddings, rather than directly mapping between different embedding dimensions. Since the adapter's output is always a probability distribution over a fixed vocabulary, it can be used to weigh and combine the corresponding vocabulary embeddings of any target LLM, regardless of their embedding dimensions.

Extensive experiments across multiple benchmarks and LLMs demonstrate the effectiveness of our approach. A key advantage of \ModelName is its ability to enable pre-training-free transfer between different LLMs. Specifically, \ModelName trained on smaller models (e.g., Qwen2-0.5B or LLaMA-3-8B) can be directly applied to larger models like Qwen2-7B and Qwen2.5-7B during supervised fine-tuning while maintaining comparable performance. This capability significantly reduces computational costs by eliminating the need for repeated pre-training when switching between different LLMs. Our key contributions can be summarized as follows:

\begin{enumerate}
\item We investigate the working principles of MLP adapters, with a particular focus on the semantic meaning of visual embeddings and how MLPs learn cross-modal alignment during training.

\item \ModelName: A Novel Adapter for LVLMs. We introduce \ModelName, a novel adapter that transforms the visual features into visual embeddings by decomposing them into linear combinations of the LLM’s vocabulary text embeddings.

\item \ModelName enables pre-training-free reuse between different large language models, significantly reducing the cost of pretraining.
\end{enumerate}



\section{Related Work}
\label{sec:related}
\subsection{Visual encoders in LVLMs}
In recent years, large language models (LLMs) \citep{zhao2023survey,minaee2024large,naveed2023comprehensive} have achieved remarkable advances in natural language processing. However, their inability to directly process visual information limits their effectiveness in multimodal tasks. To overcome this limitation, visual encoders \citep{radford2021learning,zhai2023sigmoid,sun2023eva,oquab2023dinov2,rombach2022high} have been integrated, serving as the ``eyes" of large vision-language models to leverage LLMs' extensive knowledge \citep{yu2023kola,niu2025wise,wu2025lanp} and strong instruction-following capabilities. Visual models that learn representations using only visual signals, such as DINOv2 \citep{oquab2023dinov2} and diffusion models \citep{rombach2022high}, often face challenges in achieving high performance, while models like CLIP \citep{radford2021learning} and SigLIP \citep{zhai2023sigmoid} align visual and text modalities effectively by training on extensive image-text datasets using contrastive learning. Recent research \citep{yang2024law,yao2024dense,tong2024eyes,tong2024cambrian} has also advocated for combining multiple visual encoders to better capture complex visual information. Nonetheless, aligning visual encoders with large language models remains challenging \citep{yin2024sea} due to differences in their representational spaces.
\subsection{Connectors in LVLMs}
To merge the visual extraction strengths of visual encoders with LLMs' broad knowledge and instruction-following skills, researchers have explored methods to connect these components \citep{zhu2025connector,wang2024pargo,yin2023survey,song2023bridge,zhang2024mm,gao2023llama,han2024onellm,lu2024ovis,chen2024ict,zheng2025unicode,chen2024internvl,bai2023qwen,huang2024deciphering,lin2023video,zhu2024llmbind}. Pioneering work represented by models such as Flamingo \cite{alayrac2022flamingo}, which uses cross-attention mechanisms, and BLIP-2 \cite{li2023blip}, which introduces the Q-former architecture, marks significant progress in this field. Subsequently, LLaVA \cite{liu2024visual,liu2024improved} adopted a two-stage learning approach, using a multilayer perceptron to map visual information into language embedding space, establishing itself as a most widely applicable alignment method \cite{zhu2023minigpt,lu2024deepseek,li2024mini,chen2024far}. Departing from previous approaches, we propose an innovative mapping technique. Rather than mapping visual information directly to language embedding space, we introduce ``language basis vectors" to learn linear combination weights of these basis vectors, creating a new way to represent visual information. 

\section{How LVLMs process the visual information}
In this section, we first introduce the preliminaries of current LVLMs (Sec~\ref{sec:preliminaries}). Then we conduct a comprehensive analysis of how MLPs bridge the modality gap through two aspects: 1. investigating how visual semantics are encoded in transformed embeddings (Sec~\ref{sec:sementic_in_vision_token}), and 2. exploring how MLPs learn cross-modal alignment during training (Sec~\ref{sec:mlp_learn}). Based on these insights, we propose the Language Basis Vector Projection Method (Sec~\ref{sec:Language Basis}).

\subsection{Preliminaries}
\label{sec:preliminaries}
As illustrated in Figure~\ref{fig:comparision} (a), mainstream LVLMs follow LLaVA's~\cite{liu2024improved} architecture with three key components: (1) Visual Encoder: Vision Transformers (ViTs) that extract visual features from input images; (2) Connector: Multilayer Perceptrons (MLPs) that transform these visual features into the LLM's embedding space; and (3) Large Language Model (LLM): The language model that processes the aligned embeddings and generates responses.

\subsection{Semantic Analysis of Visual Embeddings}
\label{sec:sementic_in_vision_token}
To understand how visual semantics are encoded in transformed embeddings, we analyze the relationship between visual embeddings and text embeddings through cosine similarity:

\begin{equation}
\text{Similarity}(v_i, t_k) = \frac{\mathbf{v}_i \cdot \mathbf{t}_k}{\|\mathbf{v}_i\| \|\mathbf{t}_k\|}
\end{equation}

where \(\mathbf{v}_i\) represents the visual embedding and \(\mathbf{t}_k\) denotes the LLM's vocabulary text embedding. As shown in Figure~\ref{fig:Explainable}, visual embeddings exhibit strong correlations with semantically related text tokens. For example, image patches of red apples show high similarity to tokens like ‘red' and ‘Apple'.  Moreover, some image tokens convey global information, for instance, ‘5,’ which refers to the number of apples. This result demonstrates that visual embeddings are within the LLM’s text embedding space through the MLP mapping, and are particularly close to their semantically related text embeddings.

\subsection{Cross-Modal Alignment Learning Process}
\label{sec:mlp_learn}
To investigate the development of cross-modal alignment capabilities, we conduct a temporal analysis across different training stages: Pretrain-100 steps, Pretrain-1000 steps, Pretrain-2000 steps, and final SFT. Our analysis reveals that MLPs undergo a progressive learning process in which they: 1) Initially establish basic visual-textual correspondences, 2) Gradually refine the projection of visual features into relevant text embedding subspaces, 3) Finally achieve robust semantic alignment between modalities. This evolutionary process is clearly demonstrated in Figure~\ref{fig:Explainable}, where we observe increasing semantic alignment strength and precision as training progresses. These results support our hypothesis that MLPs gradually develop their projection capabilities, incrementally learning to map visual features into regions close to the subspace spanned by their corresponding text embeddings.

\subsection{Language Basis Vector Projection Method}
\label{sec:Language Basis}
Based on our findings that MLPs learn to project visual features into regions close to the subspace spanned by their corresponding text embeddings, we propose Language Basis Vector Projection as an explicit transformation approach. This method represents visual embeddings as weighted combinations of LLM vocabulary embeddings:

\begin{equation}
\mathbf{v} = \sum_{k=1}^{N} \beta_k \mathbf{t}_k
\end{equation}

where \(\beta_k\) represents the probability distribution over vocabulary tokens. This approach offers two key advantages: 1) it explicitly models the semantic relationship between visual and textual modalities, and 2) it enables pre-training-free transfer between different LLMs since it only learns the combination weights of Language Basis Vector rather than direct dimension mapping.
\section{\ModelName}
\begin{figure*}[h]
    \centering
    \includegraphics[width=\textwidth]{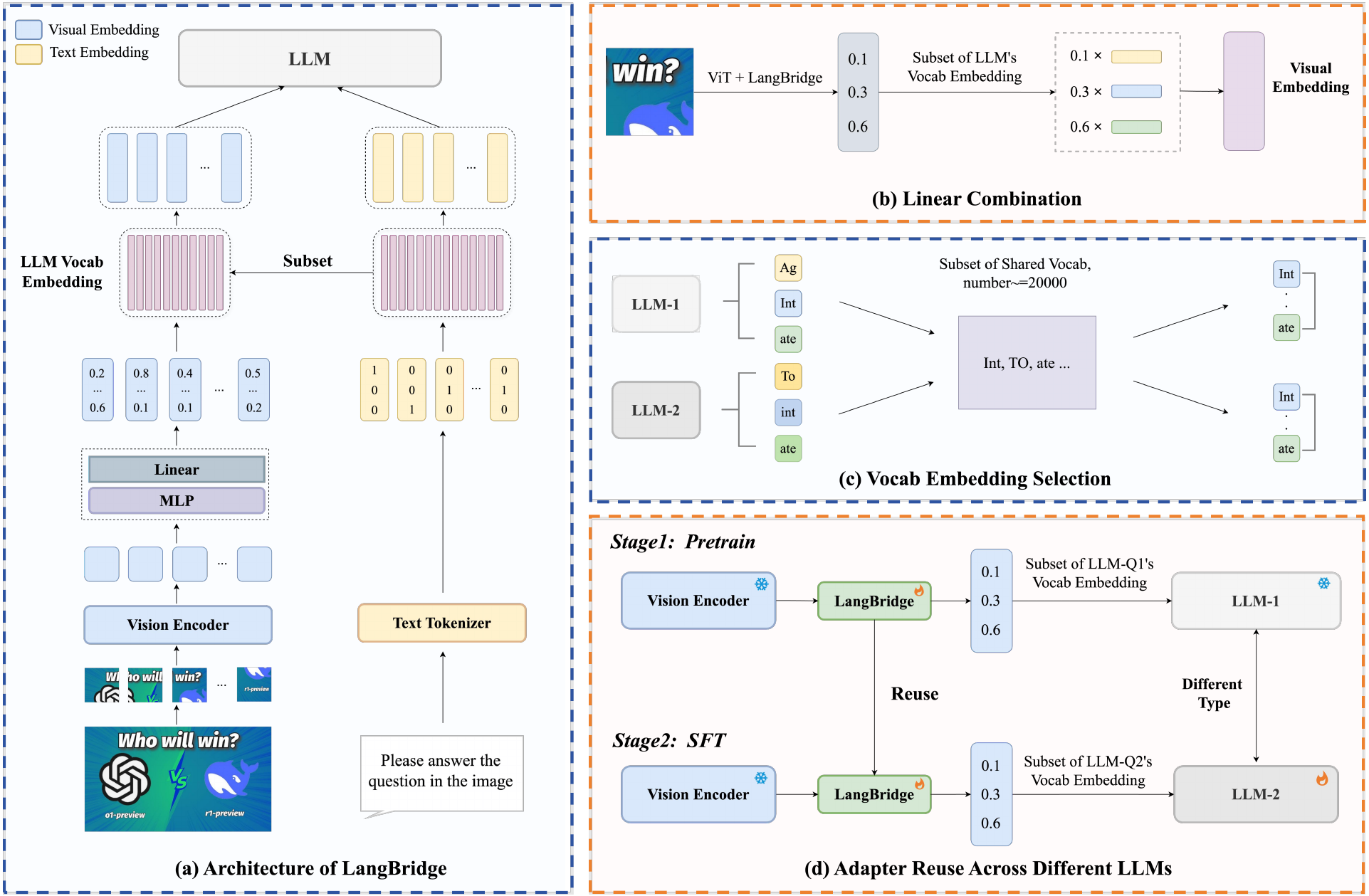}
    \vspace{-0.75em}
    \caption{\textbf{Overview of LangBridge architecture and workflow.} (a) The architecture of LangBridge:  LangBridge first extracts visual features through a Vision Encoder, then transforms them into visual embeddings by decomposing them into linear combinations of LLM's vocabulary embeddings. These visual embeddings are then concatenated with text embeddings for LLM processing. (b) Linear Combination: LangBridge generates probability distributions over the vocabulary and multiplies them with text embeddings to form visual embeddings. (c) Vocabulary Embedding Selection: A subset of shared vocabulary embeddings is selected to reduce parameter count and optimization complexity while enabling cross-LLM reuse. (d) Adapter Reuse: LangBridge shows cross-LLM adaptability, allowing an adapter pre-trained on one LLM to be reused on a different LLM during the SFT stage.}
    \vspace{-0.3mm}
    \label{fig:formulation}
\end{figure*}

In this section, we introduce \ModelName, a novel adapter that translates visual features into the language model's embedding space based on the Language Basis Vector Projection Theory. Our key insight is to decompose visual features into linear combinations of the LLM's vocabulary embeddings, enabling seamless integration with different LLM architectures. First, we present the core architecture of \ModelName (Sec~\ref{sec:LangBridge_arch}), which operates by extracting visual features from images and projecting them into probability distributions over the LLM's vocabulary space. These distributions are then used to construct visual tokens as weighted combinations of text embeddings. Second, to address the computational challenges posed by large vocabulary embedding tables (150,000 tokens), we introduce an efficient vocabulary selection strategy (Sec~\ref{sec:selectTable}). By carefully selecting a subset of approximately 20,000 most frequent tokens shared across different LLMs, we significantly reduce the parameter size and enable reuse across different LLMs. Finally, we explain how the pretrained \ModelName can be seamlessly reused across different LLM architectures without retraining (Sec~\ref{sec:transfer}).

\subsection{Architecture of \ModelName}
\label{sec:LangBridge_arch}
We present the architecture of \ModelName, a novel vision-language adapter based on the Language Basis Vector Projection Theory. As illustrated in Figure~\ref{fig:formulation}, the framework operates through three stages:

\noindent\textbf{Stage 1: Visual Feature Extraction.} Given an input image $\mathcal{I} \in \mathbb{R}^{H\times W\times 3}$, we employ a Vision Transformer to extract patch-aligned visual features:

\begin{equation}
    \{v_i\}_{i=1}^N = \text{ViT}(\mathcal{I}), \quad v_i \in \mathbb{R}^D
\end{equation}

where $N = \frac{HW}{P^2}$ denotes the number of image patches with patch size $P$.

\noindent\textbf{Stage 2: Probability Computation.} 
After acquiring the visual features, we transform them into visual tokens and align them with the LLM’s text embedding space. This process is carried out by decomposing the visual features into linear combinations of the LLM’s vocabulary embeddings. As shown in Eq.~\ref{eq:probability_computation}, we proceed as follows: We first apply a two-layer MLP to project the visual features into the LLM’s text embedding space. Let $\mathbf{v}$ represent the visual feature, The projection is given by: $\mathbf{v}_{\text{proj}} = \text{MLP}(\mathbf{v})$, where \( \mathbf{v}_{\text{proj}} \in \mathbb{R}^D \), and \( D \) is the dimensionality of the text embeddings. Next, we append a linear layer \( \mathbf{W} \in \mathbb{R}^{T \times D} \), where \( T \) is the vocabulary size of the LLM, to produce the probability distribution \( \mathbf{p} \), matching the LLM’s vocab size.

\begin{equation}
\mathbf{p} = \mathbf{W} \cdot \text{MLP}(\mathbf{v})
\label{eq:probability_computation}
\end{equation}

\noindent\textbf{Stage 3: Linear combination of text embeddings.}
Once we have the probability distribution \( \mathbf{p} \) from Stage 2, the visual features are represented as a weighted combination of the LLM’s text embeddings. The probability distribution \( \mathbf{p} \) serves as the coefficients for a linear combination of the LLM’s vocabulary embeddings \( \mathbf{e}_i \), where each \( \mathbf{e}_i \in \mathbb{R}^D \) is a text embedding of the \( i \)-th token in the vocabulary.

Specifically, the visual tokens \( \mathbf{v}_{\text{tokens}} \) are obtained as follows:

\[
\mathbf{v}_{\text{tokens}} = \sum_{i=1}^T p_i \mathbf{e}_i
\]

where \( p_i \) is the \( i \)-th component of the probability distribution \( \mathbf{p} \), and \( T \) is the vocabulary size of the LLM. The resulting vector \( \mathbf{v}_{\text{tokens}} \in \mathbb{R}^D \) represents the final visual token, which aligns with the LLM's text embedding space.

\subsection{Vocabulary Selection}
\label{sec:selectTable}
A straightforward implementation of \ModelName\ would use the entire vocabulary embedding matrix of the LLM, which can be extremely large (nearly 1B size). To mitigate these issues, we select a compact subset of $N < 20\text{k}$ tokens. To ensure the generalizability of the vocabulary and mitigate the parameter burden of \ModelName for efficient training, we merged the vocabularies of LLaMA and Qwen, retaining only the tokens present in both. Subsequently, we performed tokenization on the ShareGPT4V and  LLaVA-CC3M-Pretrain-595K datasets to generate a vocabulary frequency table. From this table, we selected the top 19,200 most frequent tokens to construct our reduced vocabulary, thereby decreasing the vocabulary size. Ultimately, the word embeddings corresponding to these 19,200 vocabulary tokens are utilized as our language basis vectors.

\subsection{Adapter Reuse Across Different LLMs}
\label{sec:transfer}
A key feature of \ModelName is its ability to be seamlessly reused across different LLM architectures after being pretrained. As shown in Figure~\ref{fig:formulation}(d), this reuse mechanism consists of two stages:

\noindent\textbf{Stage1: Pretrain.} During the pretraining stage, we first extract image features through the Vision Encoder, then use \ModelName to learn to map these features to the space of Shared Vocab Embedding of LLM-1, optimizing by the caption loss. In this stage, \ModelName outputs probability distributions over the selected vocabulary.

\noindent\textbf{Stage2: SFT.} When constructing a LVLM based on the new LLM-2, we do not need to retrain \emph{\ModelName} in stage 1. Instead, we can directly reuse the adapters that were co-trained with LLM-1 in stage 1. The core principle behind this is that \ModelName only needs to out the probability distributions over the corresponding vocab instead of direct transform. When reusing \ModelName with different LLMs, such as Qwen2-0.5B and LLaMA3-8B, the key difficulty lies in the vocabulary (due to differences in vocabulary size). To enable reuse across different LLMs, as described in Figure~\ref{fig:formulation} (c) and Section~\ref{sec:selectTable}, we construct a shared vocabulary $V_{shared}$. The reusing process can be expressed as: 

\begin{equation}
P = \text{LangBridge}_{\text{LLM}_1}(I) \in \mathbb{R}^{|V_{\text{shared}}|}
\end{equation}

\begin{equation}
\text{Visiontoken}_{\text{LLM}2} = P \cdot V_{\text{shared}}
\end{equation}

Here, \( P \) represents the probability distribution over the shared vocabulary \( V_{\text{shared}} \). The operation \( P \cdot V_{\text{shared}} \) projects this distribution onto \( \text{LLM-2's vocabulary} \), resulting in the vision token \( \text{Visiontoken}_{\text{LLM-2}} \), which is then sent to \( \text{LLM-2} \) for processing.
\section{Experiments}
\label{sec:exp}

\begin{table*}[t!]
\centering
\caption{\textbf{Same-architecture transfer results.} We evaluate the transferability of \ModelName\ between LLMs of the same architecture but different parameter sizes by using pretrained and fine-tuned connectors from Qwen2-0.5B (denoted as Qwen2-0.5B-Pretrain-\ModelName\ and Qwen2-0.5B-SFT-\ModelName, respectively). \textcolor{red}{Red} and \textcolor{blue}{blue} percentages indicate performance improvement and degradation relative to baseline models, respectively.}
\label{tab:same}
\scalebox{0.8}{
\begin{tabular}{l l | l l l l l l l}
\toprule
\multirow{2}{*}{SFT-LLM} & \multirow{2}{*}{Connector} & \multicolumn{7}{c}{Benchmark Results} \\
\cmidrule(lr){3-9}
 & & GQA & TextVQA & MME & MMBench & MMVeT & POPE & SciQA \\
\midrule
\cmidrule{1-9}
Qwen2-7B & Qwen2-7B-Pretrain-MLPs & 62.92 & 57.24 & 1938 & 72.7 & 35.5 & 87.8 & 79.44 \\
\rowcolor{lightblue}
 \text{-} &  Qwen2-0.5B-Pretrain-\ModelName & 
 63.03 \textcolor{red}{\tiny(100.17\%)} & 
 57.25 \textcolor{red}{\tiny(100.02\%)} & 
 1886 \textcolor{blue}{\tiny(97.32\%)} & 
 71.7 \textcolor{blue}{\tiny(98.62\%)} & 
 34.1 \textcolor{blue}{\tiny(96.06\%)} & 
 88.2 \textcolor{red}{\tiny(100.46\%)} & 
 79.23 \textcolor{blue}{\tiny(99.74\%)} \\
\rowcolor{lightpink}
  \text{-} &   Qwen2-0.5B-SFT-\ModelName & 
 63.15 \textcolor{red}{\tiny(100.37\%)} & 
 57.34 \textcolor{red}{\tiny(100.17\%)} & 
 1904 \textcolor{blue}{\tiny(98.25\%)} & 
 71.0 \textcolor{blue}{\tiny(97.66\%)} & 
 31.6 \textcolor{blue}{\tiny(89.01\%)} & 
 88.3 \textcolor{red}{\tiny(100.57\%)} & 
 79.25 \textcolor{blue}{\tiny(99.76\%)} \\
\cmidrule{1-9}
Qwen2.5-7B & Qwen2.5-7B-Pretrain-MLPs & 62.70 & 57.83 & 1939 & 73.8 & 34.8 & 88.4 & 79.11 \\ 
\rowcolor{lightblue}
 \text{-} &  Qwen2-0.5B-Pretrain-\ModelName & 
 62.68 \textcolor{blue}{\tiny(99.97\%)} & 
 57.35 \textcolor{blue}{\tiny(99.17\%)} & 
 1871 \textcolor{blue}{\tiny(96.49\%)} & 
 72.3 \textcolor{blue}{\tiny(97.97\%)} & 
 34.7 \textcolor{red}{\tiny(99.71\%)} & 
 88.1 \textcolor{blue}{\tiny(99.66\%)} & 
 81.23 \textcolor{red}{\tiny(102.68\%)} \\
\rowcolor{lightpink}
  \text{-} &   Qwen2-0.5B-SFT-\ModelName &  
 62.69 \textcolor{blue}{\tiny(99.98\%)} & 
 57.94 \textcolor{red}{\tiny(100.19\%)} & 
 1915 \textcolor{blue}{\tiny(98.76\%)} & 
 72.7 \textcolor{blue}{\tiny(98.51\%)} & 
 35.6 \textcolor{red}{\tiny(102.30\%)} & 
 88.1 \textcolor{blue}{\tiny(99.66\%)} & 
 77.1 \textcolor{blue}{\tiny(97.46\%)} \\
\cmidrule{1-9}
Qwen2.5-14B & Qwen2.5-14B-Pretrain-MLPs & 63.71 & 61.32 & 2038 &78.2 & 37.7 & 88.1 & 85.59 \\
\rowcolor{lightblue}
 \text{-} &  Qwen2-0.5B-Pretrain-\ModelName & 
 63.75 \textcolor{red}{\tiny(100.06\%)} & 
 61.57 \textcolor{red}{\tiny(100.41\%)} & 
 1963 \textcolor{blue}{\tiny(96.32\%)} & 
 76.2 \textcolor{blue}{\tiny(97.44\%)} & 
 35.9 \textcolor{blue}{\tiny(95.23\%)} & 
 88.2 \textcolor{red}{\tiny(100.11\%)} & 
 84.74 \textcolor{blue}{\tiny(99.01\%)} \\
\rowcolor{lightpink}
  \text{-} &   Qwen2-0.5B-SFT-\ModelName & 
 63.92 \textcolor{red}{\tiny(100.33\%)} & 
 62.02 \textcolor{red}{\tiny(101.14\%)} & 
 1990 \textcolor{blue}{\tiny(97.64\%)} & 
 77.4 \textcolor{blue}{\tiny(98.98\%)} & 
 38.4 \textcolor{red}{\tiny(101.86\%)} & 
 87.6 \textcolor{blue}{\tiny(99.43\%)} & 
 84.77 \textcolor{blue}{\tiny(99.04\%)} \\
\bottomrule
\end{tabular}
}
\end{table*}

\begin{table*}[t!]
\centering
\caption{\textbf{Cross-architecture transfer results.} We evaluate the transferability of LangBridge between different LLM architectures by using pre-trained and fine-tuned connectors from LLaMA3-8B with Qwen2-7B, Qwen2.5-7B/14B, and Qwen2-0.5B with LLaMA3-8B.}
\label{tab:cross}
\scalebox{0.8}{
\begin{tabular}{l l | l l l l l l l}
\toprule
\multirow{2}{*}{SFT-LLM} & \multirow{2}{*}{Connector} & \multicolumn{7}{c}{Benchmark Results} \\
\cmidrule(lr){3-9}
 & & GQA & TextVQA & MME & MMBench & MMVeT & POPE & SciQA \\
\midrule
\cmidrule{1-9}
LLaMA3-8B & LLaMA3-8B-Pretrain-MLPs & 63.24 & 55.66 & 1736 & \textbf{71.0} & 31.0 & 87.1 & 78.61 \\
\rowcolor{lightblue}
 \text{-} &  Qwen2-0.5B-Pretrain-\ModelName & 
 \textbf{64.14} \textcolor{red}{\tiny(101.42\%)} & 
 56.70 \textcolor{red}{\tiny(101.87\%)} & 
 1772 \textcolor{red}{\tiny(102.07\%)} & 
 70.1 \textcolor{blue}{\tiny(98.73\%)} & 
 \textbf{34.0} \textcolor{red}{\tiny(109.68\%)} & 
 \textbf{87.5} \textcolor{red}{\tiny(100.46\%)} & 
\textbf{78.78} \textcolor{red}{\tiny(100.22\%)} \\
 \rowcolor{lightpink}
  \text{-} &   Qwen2-0.5B-SFT-\ModelName & 
 63.96 \textcolor{red}{\tiny(101.14\%)} & 
 \textbf{57.35} \textcolor{red}{\tiny(103.04\%)} & 
 \textbf{1804} \textcolor{red}{\tiny(103.92\%)} & 
 69.4 \textcolor{blue}{\tiny(97.75\%)} & 
 32.7 \textcolor{red}{\tiny(105.48\%)} & 
 \textbf{87.5} \textcolor{red}{\tiny(100.46\%)} & 
 78.33 \textcolor{blue}{\tiny(99.64\%)} \\
\cmidrule{1-9}
Qwen2-7B & Qwen2-7B-Pretrain-MLPs & \textbf{62.92} & \textbf{57.24} & \textbf{1938} & \textbf{72.7} & \textbf{35.5} & 87.8 & 79.44 \\
 \rowcolor{lightblue}
 \text{-}& LLaMA3-8B-Pretrain-\ModelName & 
 62.90 \textcolor{blue}{\tiny(99.97\%)} & 
 57.23 \textcolor{blue}{\tiny(99.98\%)} & 
 1874 \textcolor{blue}{\tiny(96.70\%)} & 
 72.1 \textcolor{blue}{\tiny(99.17\%)} & 
 34.5 \textcolor{blue}{\tiny(97.18\%)} & 
 87.6 \textcolor{blue}{\tiny(99.77\%)} & 
 \textbf{80.17} \textcolor{red}{\tiny(100.92\%)} \\
 \rowcolor{lightpink}
 \text{-}& LLaMA3-8B-SFT-\ModelName & 
 62.77 \textcolor{blue}{\tiny(99.76\%)} & 
 57.08 \textcolor{blue}{\tiny(99.72\%)} & 
 1915 \textcolor{blue}{\tiny(98.81\%)} & 
 71.7 \textcolor{blue}{\tiny(98.62\%)} & 
 33.2 \textcolor{blue}{\tiny(93.52\%)} & 
 \textbf{88.2} \textcolor{red}{\tiny(100.46\%)} & 
 78.83 \textcolor{blue}{\tiny(99.23\%)} \\
\cmidrule{1-9}
Qwen2.5-7B & Qwen2.5-7B-Pretrain-MLPs & 62.70 & \textbf{57.83} & \textbf{1939} & \textbf{73.8} & 34.8 & 88.4 & 79.11 \\
 \rowcolor{lightblue}
  \text{-}& LLaMA3-8B-Pretrain-\ModelName & 
 62.72 \textcolor{red}{\tiny(100.03\%)} & 
 57.78 \textcolor{blue}{\tiny(99.91\%)} & 
 1878 \textcolor{blue}{\tiny(96.85\%)} & 
 71.9 \textcolor{blue}{\tiny(97.43\%)} & 
 32.3 \textcolor{blue}{\tiny(92.82\%)} & 
 88.4 \textcolor{red}{\tiny(100.00\%)} & 
 \textbf{80.59} \textcolor{red}{\tiny(101.87\%)} \\
\rowcolor{lightpink}
 \text{-}& LLaMA3-8B-SFT-\ModelName & 
 \textbf{63.03} \textcolor{red}{\tiny(100.53\%)} & 
 57.72 \textcolor{blue}{\tiny(99.81\%)} & 
 1909 \textcolor{blue}{\tiny(98.45\%)} & 
 71.1 \textcolor{blue}{\tiny(96.34\%)} & 
 \textbf{37.7} \textcolor{red}{\tiny(108.33\%)} & 
 \textbf{90.1} \textcolor{red}{\tiny(101.92\%)} & 
 79.84 \textcolor{red}{\tiny(100.92\%)} \\
\cmidrule{1-9}
Qwen2.5-14B & Qwen2.5-14B-Pretrain-MLPs & 63.71 & 61.32 & \textbf{2038} & \textbf{78.2} & 37.7 & \textbf{88.1} & \textbf{85.59} \\
 \rowcolor{lightblue}
 \text{-}& LLaMA3-8B-Pretrain-\ModelName & 
 63.88 \textcolor{red}{\tiny(100.27\%)} & 
 \textbf{61.39} \textcolor{blue}{\tiny(99.89\%)} & 
 2005 \textcolor{blue}{\tiny(98.38\%)} & 
 77.1 \textcolor{blue}{\tiny(98.59\%)} & 
 36.9 \textcolor{blue}{\tiny(97.88\%)} & 
 87.9 \textcolor{blue}{\tiny(99.77\%)} & 
 85.03 \textcolor{blue}{\tiny(99.35\%)} \\
  \rowcolor{lightpink}
 \text{-}& LLaMA3-8B-SFT-\ModelName & 
 \textbf{64.47} \textcolor{red}{\tiny(101.19\%)} & 
 61.35 \textcolor{blue}{\tiny(99.95\%)} & 
 1980 \textcolor{blue}{\tiny(97.15\%)} & 
 76.3 \textcolor{blue}{\tiny(97.57\%)} & 
 \textbf{38.3} \textcolor{red}{\tiny(101.59\%)} & 
 87.8 \textcolor{blue}{\tiny(99.66\%)} & 
 84.23 \textcolor{blue}{\tiny(98.41\%)} \\
\bottomrule
\end{tabular}
}
\end{table*}

\begin{table*}[t!]
\centering
\caption{\textbf{Performance comparison of different paradigms.} We compare our LangBridge with standard MLPs across seven widely used benchmarks. The results demonstrate that LangBridge achieves performance comparable to that of the standard MLP adapter.}
\scalebox{0.95}{
\begin{tabular}{l l | l l l l l l l}
\toprule
\multirow{2}{*}{LLM} & \multirow{2}{*}{Connector} & \multicolumn{7}{c}{Benchmark Results} \\
\cmidrule(lr){3-9}
 & & GQA & TextVQA & MME & MMBench & MMVeT & POPE & SciQA \\
\midrule
LLaMA3-8B & MLPs & 63.24 & 55.66 & 1736 & 71.0 & 31.0 & 87.1 & \textbf{78.61} \\
\rowcolor{lightblue}
 \text{-} & \ModelName & 
 \textbf{64.00} \textcolor{red}{\tiny(101.20\%)} & 
 \textbf{56.55} \textcolor{red}{\tiny(101.60\%)} & 
 \textbf{1775} \textcolor{red}{\tiny(102.25\%)} & 
 \textbf{72.6} \textcolor{red}{\tiny(102.25\%)} & 
 \textbf{31.1} \textcolor{red}{\tiny(100.32\%)} & 
 \textbf{87.7} \textcolor{red}{\tiny(100.69\%)} & 
 77.98 \textcolor{blue}{\tiny(99.20\%)} \\
\cmidrule{1-9}
Qwen2.5-7B & MLPs & 62.70 & \textbf{57.83} & \textbf{1939} & \textbf{73.8} & 34.8 & \textbf{88.4} & 79.11 \\
\rowcolor{lightpink}
 \text{-} & \ModelName & 
 \textbf{62.88} \textcolor{red}{\tiny(100.29\%)} & 
 57.81 \textcolor{blue}{\tiny(99.97\%)} & 
 1874 \textcolor{blue}{\tiny(96.65\%)} & 
 \textbf{73.8} \textcolor{red}{\tiny(100.00\%)} & 
 \textbf{36.7} \textcolor{red}{\tiny(105.46\%)} & 
 88.3 \textcolor{blue}{\tiny(99.89\%)} & 
 \textbf{79.18} \textcolor{red}{\tiny(100.09\%)} \\
\bottomrule
\end{tabular}
}
\label{tab:main_results}
\end{table*}

In this section, we first introduce the experimental setup of this study in detail. Subsequently, we systematically compare the performance of the proposed \ModelName with baseline models across multiple evaluation metrics and highlight its unique advantage of enabling pre-training-free transfer between different large language models. Specifically, we explore two scenarios: same-architecture transfer and cross-architecture transfer, to demonstrate the unique advantage of the proposed paradigm fully. As a novel training paradigm, we also compare \ModelName's performance with LLaVA on the same LLM (i.e., without transfer). Finally, we conduct an in-depth ablation study and provide a detailed analysis of the obtained results.

\subsection{Experiment setup}
\subsubsection{Implementation Details.}
\label{sec:Implementation Details}
Our architecture is composed of four key components: language model, visual encoder, connector, and language base vector. We selected LLaMA3-8B-Instruct \citep{dubey2024llama}, Qwen2-0.5B/7B-Instruct \citep{yang2024qwen2} and Qwen2.5-7B/14B-Instruct \citep{yang2024qwen2.5} as our language models. For brevity, we omit the ``-Instruct" suffix in the remainder of this paper. We employed the most popular CLIP-ViT-L/14@336px\citep{radford2021learning} for the visual encoder. The connector is based on an MLP architecture with a hidden size of 5120. Regarding the language base vector, we carefully selected 19,200 high-frequency language tokens. Following the LLaVA \citep{liu2024improved} setup, we utilized LLaVA-CC3M-Pretrain-595K and LLaVA-v1.5-mix665k as the pretraining and instruction-tuning datasets, respectively, training for one epoch. Specific hyperparameter settings can be found in Appendix.
\subsubsection{Benchmarks.}
We evaluated our method on seven widely-used benchmarks: GQA~\cite{hudson2019gqa}, TextVQA~\cite{singh2019towards}, MME~\cite{yin2023survey}, MMBench~\cite{liu2024mmbench}, MMVeT~\cite{yu2023mm}, POPE~\cite{li2023evaluating}, SciQA~\cite{saikh2022scienceqa}, covering a diverse range of vision-language understanding tasks. More details are in Appendix.

\subsection{Experiment results}

\subsubsection{Same-architecture transfer} 
We first examine the effectiveness of \ModelName\ in the context of same-architecture transfer. This setting is of great significance for efficiently scaling up model size, as it allows leveraging knowledge learned on smaller, more readily trainable models to improve the performance of larger ones. As shown in Table \ref{tab:same}, the \ModelName\ module, pre-trained or fine-tuned on a Qwen2-0.5B model, can be seamlessly transferred to Qwen2-7B and Qwen2.5-7B/14B models without requiring computationally expensive re-pretraining.
The results demonstrate that this transfer not only maintains strong performance but also, in several instances, exceeds the performance of the baseline models trained directly on the target LLM size. For example, when transferring the Qwen2-0.5B-SFT-\ModelName\ connector to Qwen2.5-14B, we observe improvements in TextVQA (+1.14\%) and MMVeT (+1.86\%) compared to the baseline. Similarly, the Qwen2-0.5B-Pretrain-\ModelName connector improves SciQA compared to the Qwen2-7B baseline (+2.68\%). These improvements, while modest, highlight \ModelName's ability to effectively leverage and transfer learned representations, leading to enhanced performance even without further pre-training on the larger models.

\subsubsection{Cross-architecture transfer}

\begin{table*}[t!]
\centering
\caption{\textbf{Ablation study on hidden sizes.} We compare different hidden sizes (5120 vs 1024) for LangBridge across various model configurations. Results show that our approach maintains strong performance even with reduced hidden sizes. }
\label{tab:ablation_dim}
\scalebox{0.85}{
\begin{tabular}{l l l | l l l l l l l}
\toprule
\multirow{2}{*}{LLM} & \multirow{2}{*}{Connector} & \multirow{2}{*}{Hidden Size} & \multicolumn{7}{c}{Benchmark Results} \\
\cmidrule(lr){4-10}
 & & & GQA & TextVQA & MME & MMBench & MMVeT & POPE & SciQA \\
\midrule
LLaMA3-8B & MLPs & 4096 & 63.24 & 55.66 & 1736 & 71.0 & 31.0 & 87.1 & \textbf{78.61} \\
\rowcolor{lightblue}
 \text{-} & \ModelName & \textbf{5120} & 
 \textbf{64.00} \textcolor{red}{\tiny(101.20\%)} & 
 \textbf{56.55} \textcolor{red}{\tiny(101.60\%)} & 
 \textbf{1775} \textcolor{red}{\tiny(102.25\%)} & 
 \textbf{72.6} \textcolor{red}{\tiny(102.25\%)} & 
 31.1 \textcolor{red}{\tiny(100.32\%)} & 
 \textbf{87.7} \textcolor{red}{\tiny(100.69\%)} & 
 77.98 \textcolor{blue}{\tiny(99.20\%)} \\
\rowcolor{lightpink}
 \text{-} & \ModelName & 1024 & 
 63.73 \textcolor{red}{\tiny(100.78\%)} & 
 56.71 \textcolor{red}{\tiny(101.89\%)} & 
 1773 \textcolor{red}{\tiny(102.13\%)} & 
 70.5 \textcolor{blue}{\tiny(99.30\%)} & 
 \textbf{33.8} \textcolor{red}{\tiny(109.03\%)} & 
 87.4 \textcolor{red}{\tiny(100.34\%)} & 
 78.26 \textcolor{blue}{\tiny(99.55\%)} \\
\bottomrule
\end{tabular}
}
\end{table*}

\begin{table*}[t!]
\centering
\caption{\textbf{Ablation study on vocabulary sizes.} We conduct experiments with different vocabulary sizes (19,200, 25,600, and 32,000) for LangBridge. Results demonstrate that a vocabulary size of 19,200 achieves optimal performance while maintaining minimal parameter count.}
\label{tab:ablation_vocab}
\scalebox{0.75}{
\begin{tabular}{l l l | l l l l l l l}
\toprule
\multirow{2}{*}{LLM} & \multirow{2}{*}{Connector} & \multirow{2}{*}{Vocab Size} & \multicolumn{7}{c}{Benchmark Results} \\
\cmidrule(lr){4-10}
 & & & GQA & TextVQA & MME & MMBench & MMVeT & POPE & SciQA \\
\midrule
Qwen2-7B & Qwen2-0.5B-SFT- & 19200 & 63.15 & 57.34 & 1904 & 71.0 & 31.6 & 88.3 & 79.25 \\
\rowcolor{lightblue}
 \text{-} \text{-} &  Qwen2-0.5B-SFT-\ModelName & 25600 & 
 63.13 \textcolor{blue}{\tiny(99.97\%)} & 
 57.58 \textcolor{red}{\tiny(100.42\%)} & 
 1842 \textcolor{blue}{\tiny(96.74\%)} & 
 71.8 \textcolor{red}{\tiny(101.13\%)} & 
 32.9 \textcolor{red}{\tiny(104.11\%)} & 
 87.9 \textcolor{blue}{\tiny(99.55\%)} & 
 79.01 \textcolor{blue}{\tiny(99.70\%)} \\
\rowcolor{lightpink}
 \text{-} \text{-} &  Qwen2-0.5B-SFT-\ModelName & 32000 & 
 63.11 \textcolor{blue}{\tiny(99.94\%)} & 
 57.19 \textcolor{blue}{\tiny(99.74\%)} & 
 1832 \textcolor{blue}{\tiny(96.22\%)} & 
 72.7 \textcolor{red}{\tiny(102.39\%)} & 
 33.2 \textcolor{red}{\tiny(105.06\%)} & 
 88.6 \textcolor{red}{\tiny(100.34\%)} & 
 79.11 \textcolor{blue}{\tiny(99.82\%)} \\
\bottomrule
\end{tabular}
}
\end{table*}

Furthermore, we explore whether LangBridge demonstrates excellent performance in cross-architecture transfer scenarios (e.g., Qwen, LLaMA), specifically examining transfers from Qwen series models to LLaMA3 and vice versa. As shown in Table \ref{tab:cross}, \ModelName\ demonstrates a remarkable ability to maintain strong performance, retaining a high percentage of the baseline performance, exceeding 99\% on many benchmarks. Moreover, the transfer from Qwen2-0.5B to Llama3-8B results in significant improvements, with performance gains observed across most benchmarks, especially MMVeT. 
In the reverse direction, transferring from LLaMA3-8B to Qwen models (7B to 14B), \ModelName\ maintains robust performance, consistently achieving above 99\% of the baseline scores across most benchmarks. In several cases, it even surpasses the baseline performance, such as when transferring to Qwen2.5-7B, where the LLaMA3-8B-SFT-\ModelName\ connector improves MMVeT (+8.33\%) and POPE (+1.92\%). Similarly, when applied to Qwen2.5-14B, the connector achieves gains in GQA (+1.19\%) and MMVeT (+1.59\%). These results demonstrate \ModelName's strong capability for cross-architecture transfer, effectively bridging different model families while maintaining or even enhancing performance across diverse multimodal tasks.

\subsubsection{Results of normal setting}

Having demonstrated \ModelName's unique capabilities in same-architecture transfer and cross-architecture transfer, we now turn to its direct performance as a multimodal training paradigm compared to the standard MLP approach used in LLaVA. Table \ref{tab:main_results} demonstrates that \ModelName\ achieves performance comparable to the standard MLP adapter across a range of benchmarks.  Notably, \ModelName\ consistently matches or exceeds the performance of MLPs on LLaMA3-8B for most metrics (GQA, TextVQA, MME, MMBench, MMVeT, and POPE, with SciQA slightly degrades). Furthermore, competitive results are also achieved on the smaller Qwen2.5-7B model, with improvements observed particularly in MMVeT.

These results indicate that \ModelName\ is not only an effective transfer learning technique but also a competitive alternative to standard MLP-based training paradigms for multimodal large language models.


\subsubsection{Ablation Study}

To thoroughly validate the effectiveness of our \ModelName across different architectural configurations, we first conduct ablation studies on the MLP hidden size, followed by an analysis of different vocab size.

\noindent\textbf{The effect of hidden size.}\quad
While our main experiments used a fixed hidden size of 5120 as described in Sec~\ref{sec:Implementation Details}, we performed additional experiments with a reduced hidden size of 1024 to investigate the impact of model capacity. As shown in Table~\ref{tab:ablation_dim}, when reducing the hidden size from 5120 to 1024, our \ModelName still maintains strong performance across different benchmarks. Specifically, LLaMA3-8B-LangBridge-1024 achieves gains on GQA (+0.78\%), TextVQA (+1.89\%), MME (+2.13\%), MMVeT (+9.03\%), and POPE (+0.34\%) compared with the LLaMA3-8B-MLPs. This demonstrates the effectiveness of our \ModelName across different hidden sizes.

\noindent\textbf{The effect of vocab size.}\quad
Table~\ref{tab:ablation_vocab} compares LangBridge's performance with different vocabulary sizes. While the 25,600-vocab version achieves marginal improvements on 3 benchmarks (TextVQA, MMBench, MMVeT) and the 32,000-vocab version also shows slight gains on 3 metrics, both exhibit notable declines on multimodal benchmarks like MME (-6.22\%/-7.56\%). Given that the 19,200-vocab configuration maintains optimal overall performance with minimal parameters, we select it as our final design.
\section{Conclusions}

In this work, we explore how MLPs in LVLMs bridge the vision-language modality gap. Our investigation reveals that MLPs progressively learn to project visual embeddings into subspaces spanned by their corresponding text embeddings. Drawing inspiration from this insight, we propose \ModelName, a novel adapter that translates visual features by decomposing them into linear combinations of LLM's vocabulary embeddings. This design effectively tackles a key challenge: the need for adapter retraining when switching between different LLM backbones. Extensive experiments show that \ModelName - an adapter pretrained on Qwen2-0.5B can be applied to larger models like LLaMA3-8B or Qwen2.5-14B maintaining competitive performance.

\clearpage

\maketitlesupplementary
\setcounter{section}{0}
\setcounter{table}{0}
\setcounter{figure}{0}

\renewcommand{\thesection}{S-\Alph{section}}   
\renewcommand {\thetable} {S-\arabic{table}}
\renewcommand {\thefigure} {S-\arabic{figure}}
\pagebreak

\section*{Overview}
\noindent In this supplementary material, we present more dataset details and more experimental results that are not included in the main paper. The contents include:

\begin{itemize}
    \item A comprehensive introduction to evaluation benchmarks~\ref{sec:eval_bench}.
    \item Detailed training configurations and hyperparameters~\ref{sec:training_details}.
    \item Additional experimental results on a broader range of benchmarks for LLaVA-Next integrated with LangBridge.~\ref{sec:llava_next_result}
    \item Discussion of Computational Cost.~\ref{sec:computation_cost}
    \item Discussion of current limitations and future directions~\ref{sec:limitations}.
\end{itemize}
\section{Evaluation Benchmarks}
\label{sec:eval_bench}
We evaluated our method on seven widely-used benchmarks, covering a diverse range of vision-language understanding tasks.

\begin{itemize}
    \item \textbf{GQA~\cite{hudson2019gqa}}: Evaluates the model's visual perception ability through open-ended questions.
    \item \textbf{TextVQA~\cite{singh2019towards}}: Tests the model's ability to read and reason about text in images to answer questions, focusing on text-based visual reasoning.
    \item \textbf{ScienceQA~\cite{saikh2022scienceqa}}: Provides a set of multiple-choice science questions with images to test the model's zero-shot generalization ability in scientific question answering.
    \item \textbf{MME~\cite{yin2023survey}}: A comprehensive evaluation of LVLMs across ten perception tasks (e.g., OCR and object recognition) and four cognitive tasks (e.g., commonsense reasoning, numerical computation, translation, and code reasoning).
    \item \textbf{MMBench~\cite{liu2024mmbench}}: A bilingual benchmark for evaluating LVLM's multimodal understanding capabilities, consisting of approximately 3000 multiple-choice questions covering 20 ability dimensions. The Chinese version is called MMBench-CN.
    \item \textbf{MMVeT~\cite{yu2023mm}}: A challenging multimodal benchmark designed to evaluate vision-language models' robustness and reliability. It focuses on testing fine-grained visual understanding, complex reasoning, and real-world application scenarios.
    \item \textbf{POPE~\cite{li2023evaluating}}: Evaluates the model's ability to identify specific objects in images, aiming to detect object-level hallucinations. It uses "yes/no" questions based on object annotations, with 50\% of queries targeting existing objects and 50\% targeting non-existing objects, employing random, popular, and adversarial sampling strategies.
\end{itemize}

Through these comprehensive benchmarks, we systematically evaluated the model's capabilities across diverse tasks, with particular emphasis on multimodal understanding, visual reasoning, hallucination detection, and real-world applicability. These benchmarks collectively provide a thorough assessment of the model's strengths and potential areas for improvement.

\section{Training Details}
\label{sec:training_details}
As shown in Table~\ref{tab:training_details}, we follow the same training recipes as LLaVA’s settings~\cite{liu2024visual} for standard MLPs, except that we change the pretrain learning rate from 1e-3 to 2e-5 for LangBridge.

\begin{table}[h]
\centering
\caption{Training hyper-parameters}
\vspace{\baselineskip}
\label{tab:training_details}
\small
\begin{tabular}{@{}l|c@{}}
\toprule
Hyper-parameter & Value \\ \midrule
batch size & 256 (pretrain), 128 (finetune) \\
learning rate & 1e-3 (pretrain), 2e-5 (finetune) \\
learning rate schedule & cosine \\
learning rate warm-up ratio & 0.03 \\
weight decay & 0 \\
epoch & 1 \\
optimizer & AdamW \\
float precision & bfloat16 \\
deepspeed configuration & zero2 (pretrain), zero3 (finetune) \\ \bottomrule
\end{tabular}
\end{table}

\section{More experiments results}
\label{sec:llava_next_result}
To further evaluate the generalizability and robustness of our method, we integrate it into LLaVA-Next. Our model first uses a Qwen2-0.5B pretrained LangBridge module, which is then combined with Qwen2-7B for further supervised fine-tuning (SFT). We compare it to a standard MLP baseline with same backbone and evaluate across a range of benchmarks. As shown in Table~\ref{tab:finegrained_benchmarks}, our model achieves consistent improvements over the baseline in fine-grained image analysis tasks, including AI2D, ChartQA, and DocVQA, while maintaining parity on MMVP. For grounding and real-world reasoning tasks (Table~\ref{tab:grounding_realworld}), our approach significantly outperforms the baseline on RefCOCO (+17.06\%) and RefCOCO+ (+15.23\%), and performs comparably on RefCOCOg and MMRealWorld. In the video understanding domain (Table~\ref{tab:video_benchmarks}), the model improves results on Seed-Video and Seed2-Video while maintaining similar performance on VideoMME and MMT. Lastly, as shown in Table~\ref{tab:common_datasets}, our method improves accuracy on TextVQA, GQA, and MME, demonstrating its effectiveness on widely used benchmarks, with only a slight drop on MMMU. These results collectively demonstrate that LangBridge consistently matches or outperforms standard MLPs across a wide range of tasks, showcasing strong generalization capability.

\begin{table}[h]
\centering
\vspace{-0.3cm}
\caption{Results on Fine-grained Benchmarks}
\vspace{-0.3cm}
\label{tab:finegrained_benchmarks}
\scalebox{0.75}{
\begin{tabular}{l | l l l l}
\toprule
Method & AI2D & ChartQA & DocVQA & MMVP \\
\midrule
Baseline & 0.763 & 0.8632 & 0.741 & 43.3 \\
\rowcolor{lightblue}
Our Model &
  0.766 \textcolor{red}{\tiny(100.39\%)} &
  0.877 \textcolor{red}{\tiny(101.60\%)} &
  0.758 \textcolor{red}{\tiny(102.29\%)} &
  43.3 \textcolor{black}{\tiny(100.00\%)} \\
\bottomrule
\end{tabular}
}
\end{table}

\vspace{-0.7cm}

\begin{table}[h]
\centering
\caption{Results on Grounding and Real-World Benchmarks}
\vspace{-0.3cm}
\label{tab:grounding_realworld}
\scalebox{0.7}{
\begin{tabular}{l | l l l l}
\toprule
Method & RefCOCO & RefCOCO+ & RefCOCOg & MMRealWorld \\
\midrule
Baseline & 0.387 & 0.348 & 0.6377 & 0.4405 \\
\rowcolor{lightblue}
Our Model &
  0.453 \textcolor{red}{\tiny(117.06\%)} &
  0.401 \textcolor{red}{\tiny(115.23\%)} &
  0.622 \textcolor{blue}{\tiny(97.53\%)} &
  0.4436 \textcolor{red}{\tiny(100.70\%)} \\
\bottomrule
\end{tabular}
}
\end{table}

\vspace{-0.4cm}

\begin{table}[h]
\centering
\vspace{-0.3cm}
\caption{Results on Video Benchmarks}
\vspace{-0.3cm}
\label{tab:video_benchmarks}
\scalebox{0.75}{
\begin{tabular}{l | l l l l}
\toprule
Method & VideoMME & Seed-Video & Seed2-Video & MMT \\
\midrule
Baseline & 50.77 & 0.444 & 0.494 & 54.43 \\
\rowcolor{lightblue}
Our Model &
  50.71 \textcolor{blue}{\tiny(99.88\%)} &
  0.450 \textcolor{red}{\tiny(101.35\%)} &
  0.504 \textcolor{red}{\tiny(102.03\%)} &
  53.98 \textcolor{blue}{\tiny(99.17\%)} \\
\bottomrule
\end{tabular}
}
\end{table}

\begin{table}[h]
\centering
\vspace{-0.7cm}
\caption{Results on Commonly Used Datasets}
\vspace{-0.3cm}
\label{tab:common_datasets}
\scalebox{0.75}{
\begin{tabular}{l | l l l l}
\toprule
Method & TextVQA & GQA & MME & MMMU \\
\midrule
Baseline & 0.654 & 0.651 & 1886 & 0.4356 \\
\rowcolor{lightblue}
Our Model &
  0.665 \textcolor{red}{\tiny(101.68\%)} &
  0.652 \textcolor{red}{\tiny(100.15\%)} &
  1928 \textcolor{red}{\tiny(102.23\%)} &
  0.4178 \textcolor{blue}{\tiny(95.91\%)} \\
\bottomrule
\end{tabular}
}
\end{table}

\section{Computational Cost}
\label{sec:computation_cost}

We follow LLaVA-Next settings and conduct experiments on H100 using its SFT data. As shown in the table below, LangBridge incurs only a ~10\% increase in training time compared to the baseline, while significantly reducing pre-training time.

\begin{table}[h!]

\centering

\label{tab:my_compact_table} 

\setlength{\tabcolsep}{4pt} 

\resizebox{\columnwidth}{!}{%

\begin{tabular}{@{}llcccccc@{}}

\toprule

\textbf{Vision} & \textbf{Backbone LLM} & \textbf{MLP} & \textbf{Speed (s/iter)} & \textbf{Time (h)} & \textbf{Parameters} & \textbf{Data} & \textbf{Hardware} \\

\midrule

CLIP-L/336 & Qwen2-7B-Instruct & Normal& 3.876 & 24.8 & 7.3B & LLaVA1.6 SFT & 8$\times$ H100 \\

CLIP-L/336 & Qwen2-7B-Instruct & Langbridge & 4.273 (1.1$\times$) & 27.3 & 7.43B & LLaVA1.6 SFT & 8$\times$ H100 \\

\bottomrule

\end{tabular}%

}

\end{table}
\vspace{-0.3cm}

\section{Limitations}
\label{sec:limitations}
While our work demonstrates promising results in vision-language tasks, it has a notable limitation: we only focus on the visual modality. Modern multimodal systems often need to process various types of inputs beyond images, such as videos, audio, and 3D data. Future work could explore extending LangBridge to support these additional modalities, potentially enabling a more comprehensive multimodal understanding system.
{
    \small
    \bibliographystyle{ieeenat_fullname}
    \bibliography{main}

\begin{thebibliography}{54}
\providecommand{\natexlab}[1]{#1}
\providecommand{\url}[1]{\texttt{#1}}
\expandafter\ifx\csname urlstyle\endcsname\relax
  \providecommand{\doi}[1]{doi: #1}\else
  \providecommand{\doi}{doi: \begingroup \urlstyle{rm}\Url}\fi

\bibitem[Alayrac et~al.(2022)Alayrac, Donahue, Luc, Miech, Barr, Hasson, Lenc, Mensch, Millican, Reynolds, et~al.]{alayrac2022flamingo}
Jean-Baptiste Alayrac, Jeff Donahue, Pauline Luc, Antoine Miech, Iain Barr, Yana Hasson, Karel Lenc, Arthur Mensch, Katherine Millican, Malcolm Reynolds, et~al.
\newblock Flamingo: a visual language model for few-shot learning.
\newblock \emph{Advances in neural information processing systems}, 35:\penalty0 23716--23736, 2022.

\bibitem[Bai et~al.(2023)Bai, Bai, Yang, Wang, Tan, Wang, Lin, Zhou, and Zhou]{bai2023qwen}
Jinze Bai, Shuai Bai, Shusheng Yang, Shijie Wang, Sinan Tan, Peng Wang, Junyang Lin, Chang Zhou, and Jingren Zhou.
\newblock Qwen-vl: A frontier large vision-language model with versatile abilities.
\newblock \emph{arXiv preprint arXiv:2308.12966}, 2023.

\bibitem[Brohan et~al.(2023)Brohan, Brown, Carbajal, Chebotar, Chen, Choromanski, Ding, Driess, Dubey, Finn, et~al.]{brohan2023rt}
Anthony Brohan, Noah Brown, Justice Carbajal, Yevgen Chebotar, Xi Chen, Krzysztof Choromanski, Tianli Ding, Danny Driess, Avinava Dubey, Chelsea Finn, et~al.
\newblock Rt-2: Vision-language-action models transfer web knowledge to robotic control.
\newblock \emph{arXiv preprint arXiv:2307.15818}, 2023.

\bibitem[Chen et~al.(2024{\natexlab{a}})Chen, Zhang, Huang, Niu, Zhang, Wen, and Hu]{chen2024ict}
Junzhe Chen, Tianshu Zhang, Shiyu Huang, Yuwei Niu, Linfeng Zhang, Lijie Wen, and Xuming Hu.
\newblock Ict: Image-object cross-level trusted intervention for mitigating object hallucination in large vision-language models.
\newblock \emph{arXiv preprint arXiv:2411.15268}, 2024{\natexlab{a}}.

\bibitem[Chen et~al.(2024{\natexlab{b}})Chen, Wang, Tian, Ye, Gao, Cui, Tong, Hu, Luo, Ma, et~al.]{chen2024far}
Zhe Chen, Weiyun Wang, Hao Tian, Shenglong Ye, Zhangwei Gao, Erfei Cui, Wenwen Tong, Kongzhi Hu, Jiapeng Luo, Zheng Ma, et~al.
\newblock How far are we to gpt-4v? closing the gap to commercial multimodal models with open-source suites.
\newblock \emph{arXiv preprint arXiv:2404.16821}, 2024{\natexlab{b}}.

\bibitem[Chen et~al.(2024{\natexlab{c}})Chen, Wu, Wang, Su, Chen, Xing, Zhong, Zhang, Zhu, Lu, et~al.]{chen2024internvl}
Zhe Chen, Jiannan Wu, Wenhai Wang, Weijie Su, Guo Chen, Sen Xing, Muyan Zhong, Qinglong Zhang, Xizhou Zhu, Lewei Lu, et~al.
\newblock Internvl: Scaling up vision foundation models and aligning for generic visual-linguistic tasks.
\newblock In \emph{Proceedings of the IEEE/CVF Conference on Computer Vision and Pattern Recognition}, pages 24185--24198, 2024{\natexlab{c}}.

\bibitem[Dubey et~al.(2024)Dubey, Jauhri, Pandey, Kadian, Al-Dahle, Letman, Mathur, Schelten, Yang, Fan, et~al.]{dubey2024llama}
Abhimanyu Dubey, Abhinav Jauhri, Abhinav Pandey, Abhishek Kadian, Ahmad Al-Dahle, Aiesha Letman, Akhil Mathur, Alan Schelten, Amy Yang, Angela Fan, et~al.
\newblock The llama 3 herd of models.
\newblock \emph{arXiv preprint arXiv:2407.21783}, 2024.

\bibitem[Gao et~al.(2023)Gao, Han, Zhang, Lin, Geng, Zhou, Zhang, Lu, He, Yue, et~al.]{gao2023llama}
Peng Gao, Jiaming Han, Renrui Zhang, Ziyi Lin, Shijie Geng, Aojun Zhou, Wei Zhang, Pan Lu, Conghui He, Xiangyu Yue, et~al.
\newblock Llama-adapter v2: Parameter-efficient visual instruction model.
\newblock \emph{arXiv preprint arXiv:2304.15010}, 2023.

\bibitem[Han et~al.(2024)Han, Gong, Zhang, Wang, Zhang, Lin, Qiao, Gao, and Yue]{han2024onellm}
Jiaming Han, Kaixiong Gong, Yiyuan Zhang, Jiaqi Wang, Kaipeng Zhang, Dahua Lin, Yu Qiao, Peng Gao, and Xiangyu Yue.
\newblock Onellm: One framework to align all modalities with language.
\newblock In \emph{Proceedings of the IEEE/CVF Conference on Computer Vision and Pattern Recognition}, pages 26584--26595, 2024.

\bibitem[Huang et~al.(2024)Huang, Dong, Zhang, Zang, Cao, Wang, Lin, Zhang, and Yu]{huang2024deciphering}
Qidong Huang, Xiaoyi Dong, Pan Zhang, Yuhang Zang, Yuhang Cao, Jiaqi Wang, Dahua Lin, Weiming Zhang, and Nenghai Yu.
\newblock Deciphering cross-modal alignment in large vision-language models with modality integration rate.
\newblock \emph{arXiv preprint arXiv:2410.07167}, 2024.

\bibitem[Hudson and Manning(2019)]{hudson2019gqa}
Drew~A Hudson and Christopher~D Manning.
\newblock Gqa: A new dataset for real-world visual reasoning and compositional question answering.
\newblock In \emph{Proceedings of the IEEE/CVF conference on computer vision and pattern recognition}, pages 6700--6709, 2019.

\bibitem[Kaduri et~al.(2024)Kaduri, Bagon, and Dekel]{kaduri2024s}
Omri Kaduri, Shai Bagon, and Tali Dekel.
\newblock What's in the image? a deep-dive into the vision of vision language models.
\newblock \emph{arXiv preprint arXiv:2411.17491}, 2024.

\bibitem[Li et~al.(2023{\natexlab{a}})Li, Li, Savarese, and Hoi]{li2023blip}
Junnan Li, Dongxu Li, Silvio Savarese, and Steven Hoi.
\newblock Blip-2: Bootstrapping language-image pre-training with frozen image encoders and large language models.
\newblock In \emph{International conference on machine learning}, pages 19730--19742. PMLR, 2023{\natexlab{a}}.

\bibitem[Li et~al.(2023{\natexlab{b}})Li, Du, Zhou, Wang, Zhao, and Wen]{li2023evaluating}
Yifan Li, Yifan Du, Kun Zhou, Jinpeng Wang, Wayne~Xin Zhao, and Ji-Rong Wen.
\newblock Evaluating object hallucination in large vision-language models.
\newblock \emph{arXiv preprint arXiv:2305.10355}, 2023{\natexlab{b}}.

\bibitem[Li et~al.(2024)Li, Zhang, Wang, Zhong, Chen, Chu, Liu, and Jia]{li2024mini}
Yanwei Li, Yuechen Zhang, Chengyao Wang, Zhisheng Zhong, Yixin Chen, Ruihang Chu, Shaoteng Liu, and Jiaya Jia.
\newblock Mini-gemini: Mining the potential of multi-modality vision language models.
\newblock \emph{arXiv preprint arXiv:2403.18814}, 2024.

\bibitem[Lin et~al.(2023)Lin, Ye, Zhu, Cui, Ning, Jin, and Yuan]{lin2023video}
Bin Lin, Yang Ye, Bin Zhu, Jiaxi Cui, Munan Ning, Peng Jin, and Li Yuan.
\newblock Video-llava: Learning united visual representation by alignment before projection.
\newblock \emph{arXiv preprint arXiv:2311.10122}, 2023.

\bibitem[Lin et~al.(2024)Lin, Tang, Ye, Cui, Zhu, Jin, Huang, Zhang, Pang, Ning, et~al.]{lin2024moe}
Bin Lin, Zhenyu Tang, Yang Ye, Jiaxi Cui, Bin Zhu, Peng Jin, Jinfa Huang, Junwu Zhang, Yatian Pang, Munan Ning, et~al.
\newblock Moe-llava: Mixture of experts for large vision-language models.
\newblock \emph{arXiv preprint arXiv:2401.15947}, 2024.

\bibitem[Liu et~al.(2024{\natexlab{a}})Liu, Li, Li, and Lee]{liu2024improved}
Haotian Liu, Chunyuan Li, Yuheng Li, and Yong~Jae Lee.
\newblock Improved baselines with visual instruction tuning.
\newblock In \emph{Proceedings of the IEEE/CVF Conference on Computer Vision and Pattern Recognition}, pages 26296--26306, 2024{\natexlab{a}}.

\bibitem[Liu et~al.(2024{\natexlab{b}})Liu, Li, Wu, and Lee]{liu2024visual}
Haotian Liu, Chunyuan Li, Qingyang Wu, and Yong~Jae Lee.
\newblock Visual instruction tuning.
\newblock \emph{Advances in neural information processing systems}, 36, 2024{\natexlab{b}}.

\bibitem[Liu et~al.(2024{\natexlab{c}})Liu, Duan, Zhang, Li, Zhang, Zhao, Yuan, Wang, He, Liu, et~al.]{liu2024mmbench}
Yuan Liu, Haodong Duan, Yuanhan Zhang, Bo Li, Songyang Zhang, Wangbo Zhao, Yike Yuan, Jiaqi Wang, Conghui He, Ziwei Liu, et~al.
\newblock Mmbench: Is your multi-modal model an all-around player?
\newblock In \emph{European conference on computer vision}, pages 216--233. Springer, 2024{\natexlab{c}}.

\bibitem[Lu et~al.(2024{\natexlab{a}})Lu, Liu, Zhang, Wang, Dong, Liu, Sun, Ren, Li, Yang, et~al.]{lu2024deepseek}
Haoyu Lu, Wen Liu, Bo Zhang, Bingxuan Wang, Kai Dong, Bo Liu, Jingxiang Sun, Tongzheng Ren, Zhuoshu Li, Hao Yang, et~al.
\newblock Deepseek-vl: towards real-world vision-language understanding.
\newblock \emph{arXiv preprint arXiv:2403.05525}, 2024{\natexlab{a}}.

\bibitem[Lu et~al.(2024{\natexlab{b}})Lu, Li, Chen, Xu, Luo, Zhang, and Ye]{lu2024ovis}
Shiyin Lu, Yang Li, Qing-Guo Chen, Zhao Xu, Weihua Luo, Kaifu Zhang, and Han-Jia Ye.
\newblock Ovis: Structural embedding alignment for multimodal large language model.
\newblock \emph{arXiv preprint arXiv:2405.20797}, 2024{\natexlab{b}}.

\bibitem[Minaee et~al.(2024)Minaee, Mikolov, Nikzad, Chenaghlu, Socher, Amatriain, and Gao]{minaee2024large}
Shervin Minaee, Tomas Mikolov, Narjes Nikzad, Meysam Chenaghlu, Richard Socher, Xavier Amatriain, and Jianfeng Gao.
\newblock Large language models: A survey.
\newblock \emph{arXiv preprint arXiv:2402.06196}, 2024.

\bibitem[Narayanaswamy(2024)]{narayanaswamy2024using}
Anand Narayanaswamy.
\newblock Using copilot in microsoft 365.
\newblock In \emph{Microsoft Copilot for Windows 11: Understanding the AI-Powered Features in Windows 11}, pages 205--233. Springer, 2024.

\bibitem[Naveed et~al.(2023)Naveed, Khan, Qiu, Saqib, Anwar, Usman, Akhtar, Barnes, and Mian]{naveed2023comprehensive}
Humza Naveed, Asad~Ullah Khan, Shi Qiu, Muhammad Saqib, Saeed Anwar, Muhammad Usman, Naveed Akhtar, Nick Barnes, and Ajmal Mian.
\newblock A comprehensive overview of large language models.
\newblock \emph{arXiv preprint arXiv:2307.06435}, 2023.

\bibitem[Niu et~al.(2025)Niu, Ning, Zheng, Lin, Jin, Liao, Ning, Zhu, and Yuan]{niu2025wise}
Yuwei Niu, Munan Ning, Mengren Zheng, Bin Lin, Peng Jin, Jiaqi Liao, Kunpeng Ning, Bin Zhu, and Li Yuan.
\newblock Wise: A world knowledge-informed semantic evaluation for text-to-image generation.
\newblock \emph{arXiv preprint arXiv:2503.07265}, 2025.

\bibitem[Oquab et~al.(2023)Oquab, Darcet, Moutakanni, Vo, Szafraniec, Khalidov, Fernandez, Haziza, Massa, El-Nouby, et~al.]{oquab2023dinov2}
Maxime Oquab, Timoth{\'e}e Darcet, Th{\'e}o Moutakanni, Huy Vo, Marc Szafraniec, Vasil Khalidov, Pierre Fernandez, Daniel Haziza, Francisco Massa, Alaaeldin El-Nouby, et~al.
\newblock Dinov2: Learning robust visual features without supervision.
\newblock \emph{arXiv preprint arXiv:2304.07193}, 2023.

\bibitem[Radford et~al.(2021)Radford, Kim, Hallacy, Ramesh, Goh, Agarwal, Sastry, Askell, Mishkin, Clark, et~al.]{radford2021learning}
Alec Radford, Jong~Wook Kim, Chris Hallacy, Aditya Ramesh, Gabriel Goh, Sandhini Agarwal, Girish Sastry, Amanda Askell, Pamela Mishkin, Jack Clark, et~al.
\newblock Learning transferable visual models from natural language supervision.
\newblock In \emph{International conference on machine learning}, pages 8748--8763. PMLR, 2021.

\bibitem[Rombach et~al.(2022)Rombach, Blattmann, Lorenz, Esser, and Ommer]{rombach2022high}
Robin Rombach, Andreas Blattmann, Dominik Lorenz, Patrick Esser, and Bj{\"o}rn Ommer.
\newblock High-resolution image synthesis with latent diffusion models.
\newblock In \emph{Proceedings of the IEEE/CVF conference on computer vision and pattern recognition}, pages 10684--10695, 2022.

\bibitem[Saikh et~al.(2022)Saikh, Ghosal, Mittal, Ekbal, and Bhattacharyya]{saikh2022scienceqa}
Tanik Saikh, Tirthankar Ghosal, Amish Mittal, Asif Ekbal, and Pushpak Bhattacharyya.
\newblock Scienceqa: A novel resource for question answering on scholarly articles.
\newblock \emph{International Journal on Digital Libraries}, 23\penalty0 (3):\penalty0 289--301, 2022.

\bibitem[Sima et~al.(2023)Sima, Renz, Chitta, Chen, Zhang, Xie, Luo, Geiger, and Li]{sima2023drivelm}
Chonghao Sima, Katrin Renz, Kashyap Chitta, Li Chen, Hanxue Zhang, Chengen Xie, Ping Luo, Andreas Geiger, and Hongyang Li.
\newblock Drivelm: Driving with graph visual question answering.
\newblock \emph{arXiv preprint arXiv:2312.14150}, 2023.

\bibitem[Singh et~al.(2019)Singh, Natarjan, Shah, Jiang, Chen, Parikh, and Rohrbach]{singh2019towards}
Amanpreet Singh, Vivek Natarjan, Meet Shah, Yu Jiang, Xinlei Chen, Devi Parikh, and Marcus Rohrbach.
\newblock Towards vqa models that can read.
\newblock In \emph{Proceedings of the IEEE Conference on Computer Vision and Pattern Recognition}, pages 8317--8326, 2019.

\bibitem[Song et~al.(2023)Song, Li, Li, Zhao, Yu, Ma, Mao, and Zhang]{song2023bridge}
Shezheng Song, Xiaopeng Li, Shasha Li, Shan Zhao, Jie Yu, Jun Ma, Xiaoguang Mao, and Weimin Zhang.
\newblock How to bridge the gap between modalities: A comprehensive survey on multimodal large language model.
\newblock \emph{arXiv preprint arXiv:2311.07594}, 2023.

\bibitem[Sun et~al.(2023)Sun, Fang, Wu, Wang, and Cao]{sun2023eva}
Quan Sun, Yuxin Fang, Ledell Wu, Xinlong Wang, and Yue Cao.
\newblock Eva-clip: Improved training techniques for clip at scale.
\newblock \emph{arXiv preprint arXiv:2303.15389}, 2023.

\bibitem[Tong et~al.(2024{\natexlab{a}})Tong, Brown, Wu, Woo, Middepogu, Akula, Yang, Yang, Iyer, Pan, et~al.]{tong2024cambrian}
Shengbang Tong, Ellis Brown, Penghao Wu, Sanghyun Woo, Manoj Middepogu, Sai~Charitha Akula, Jihan Yang, Shusheng Yang, Adithya Iyer, Xichen Pan, et~al.
\newblock Cambrian-1: A fully open, vision-centric exploration of multimodal llms.
\newblock \emph{arXiv preprint arXiv:2406.16860}, 2024{\natexlab{a}}.

\bibitem[Tong et~al.(2024{\natexlab{b}})Tong, Liu, Zhai, Ma, LeCun, and Xie]{tong2024eyes}
Shengbang Tong, Zhuang Liu, Yuexiang Zhai, Yi Ma, Yann LeCun, and Saining Xie.
\newblock Eyes wide shut? exploring the visual shortcomings of multimodal llms.
\newblock In \emph{Proceedings of the IEEE/CVF Conference on Computer Vision and Pattern Recognition}, pages 9568--9578, 2024{\natexlab{b}}.

\bibitem[Wang et~al.(2024)Wang, Shan, Shi, Lin, Fei, Tang, Liao, Tang, Huang, and Zheng]{wang2024pargo}
An-Lan Wang, Bin Shan, Wei Shi, Kun-Yu Lin, Xiang Fei, Guozhi Tang, Lei Liao, Jingqun Tang, Can Huang, and Wei-Shi Zheng.
\newblock Pargo: Bridging vision-language with partial and global views.
\newblock \emph{arXiv preprint arXiv:2408.12928}, 2024.

\bibitem[Wu et~al.(2025)Wu, Niu, Gao, Lin, Zhang, Zhang, Shi, Wang, Fu, Xu, et~al.]{wu2025lanp}
Zongyu Wu, Yuwei Niu, Hongcheng Gao, Minhua Lin, Zhiwei Zhang, Zhifang Zhang, Qi Shi, Yilong Wang, Sike Fu, Junjie Xu, et~al.
\newblock Lanp: Rethinking the impact of language priors in large vision-language models.
\newblock \emph{arXiv preprint arXiv:2502.12359}, 2025.

\bibitem[Yang et~al.(2024{\natexlab{a}})Yang, Yang, Hui, Zheng, Yu, Zhou, Li, Li, Liu, Huang, et~al.]{yang2024qwen2}
An Yang, Baosong Yang, Binyuan Hui, Bo Zheng, Bowen Yu, Chang Zhou, Chengpeng Li, Chengyuan Li, Dayiheng Liu, Fei Huang, et~al.
\newblock Qwen2 technical report.
\newblock \emph{arXiv preprint arXiv:2407.10671}, 2024{\natexlab{a}}.

\bibitem[Yang et~al.(2024{\natexlab{b}})Yang, Yang, Zhang, Hui, Zheng, Yu, Li, Liu, Huang, Wei, et~al.]{yang2024qwen2.5}
An Yang, Baosong Yang, Beichen Zhang, Binyuan Hui, Bo Zheng, Bowen Yu, Chengyuan Li, Dayiheng Liu, Fei Huang, Haoran Wei, et~al.
\newblock Qwen2. 5 technical report.
\newblock \emph{arXiv preprint arXiv:2412.15115}, 2024{\natexlab{b}}.

\bibitem[Yang et~al.(2024{\natexlab{c}})Yang, Zhai, You, Yuan, Yang, and Xu]{yang2024law}
Shijia Yang, Bohan Zhai, Quanzeng You, Jianbo Yuan, Hongxia Yang, and Chenfeng Xu.
\newblock Law of vision representation in mllms.
\newblock \emph{arXiv preprint arXiv:2408.16357}, 2024{\natexlab{c}}.

\bibitem[Yao et~al.(2024)Yao, Wu, Yang, Song, Zhang, Feng, Sun, Li, Ouyang, and Wang]{yao2024dense}
Huanjin Yao, Wenhao Wu, Taojiannan Yang, YuXin Song, Mengxi Zhang, Haocheng Feng, Yifan Sun, Zhiheng Li, Wanli Ouyang, and Jingdong Wang.
\newblock Dense connector for mllms.
\newblock \emph{arXiv preprint arXiv:2405.13800}, 2024.

\bibitem[Yin et~al.(2023)Yin, Fu, Zhao, Li, Sun, Xu, and Chen]{yin2023survey}
Shukang Yin, Chaoyou Fu, Sirui Zhao, Ke Li, Xing Sun, Tong Xu, and Enhong Chen.
\newblock A survey on multimodal large language models.
\newblock \emph{arXiv preprint arXiv:2306.13549}, 2023.

\bibitem[Yin et~al.(2024)Yin, Zhao, Zhang, Lin, Wang, Tao, Wan, Zhang, Yin, and Zhang]{yin2024sea}
Yuanyang Yin, Yaqi Zhao, Yajie Zhang, Ke Lin, Jiahao Wang, Xin Tao, Pengfei Wan, Di Zhang, Baoqun Yin, and Wentao Zhang.
\newblock Sea: Supervised embedding alignment for token-level visual-textual integration in mllms.
\newblock \emph{arXiv preprint arXiv:2408.11813}, 2024.

\bibitem[Yu et~al.(2023{\natexlab{a}})Yu, Wang, Tu, Cao, Zhang-Li, Lv, Peng, Yao, Zhang, Li, et~al.]{yu2023kola}
Jifan Yu, Xiaozhi Wang, Shangqing Tu, Shulin Cao, Daniel Zhang-Li, Xin Lv, Hao Peng, Zijun Yao, Xiaohan Zhang, Hanming Li, et~al.
\newblock Kola: Carefully benchmarking world knowledge of large language models.
\newblock \emph{arXiv preprint arXiv:2306.09296}, 2023{\natexlab{a}}.

\bibitem[Yu et~al.(2023{\natexlab{b}})Yu, Yang, Li, Wang, Lin, Liu, Wang, and Wang]{yu2023mm}
Weihao Yu, Zhengyuan Yang, Linjie Li, Jianfeng Wang, Kevin Lin, Zicheng Liu, Xinchao Wang, and Lijuan Wang.
\newblock Mm-vet: Evaluating large multimodal models for integrated capabilities.
\newblock \emph{arXiv preprint arXiv:2308.02490}, 2023{\natexlab{b}}.

\bibitem[Zhai et~al.(2023)Zhai, Mustafa, Kolesnikov, and Beyer]{zhai2023sigmoid}
Xiaohua Zhai, Basil Mustafa, Alexander Kolesnikov, and Lucas Beyer.
\newblock Sigmoid loss for language image pre-training.
\newblock In \emph{Proceedings of the IEEE/CVF International Conference on Computer Vision}, pages 11975--11986, 2023.

\bibitem[Zhang et~al.(2024{\natexlab{a}})Zhang, Yu, Dong, Li, Su, Chu, and Yu]{zhang2024mm}
Duzhen Zhang, Yahan Yu, Jiahua Dong, Chenxing Li, Dan Su, Chenhui Chu, and Dong Yu.
\newblock Mm-llms: Recent advances in multimodal large language models.
\newblock \emph{arXiv preprint arXiv:2401.13601}, 2024{\natexlab{a}}.

\bibitem[Zhang et~al.(2024{\natexlab{b}})Zhang, Shen, Li, and Liu]{zhang2024large}
Kaichen Zhang, Yifei Shen, Bo Li, and Ziwei Liu.
\newblock Large multi-modal models can interpret features in large multi-modal models.
\newblock \emph{arXiv preprint arXiv:2411.14982}, 2024{\natexlab{b}}.

\bibitem[Zhao et~al.(2023)Zhao, Zhou, Li, Tang, Wang, Hou, Min, Zhang, Zhang, Dong, et~al.]{zhao2023survey}
Wayne~Xin Zhao, Kun Zhou, Junyi Li, Tianyi Tang, Xiaolei Wang, Yupeng Hou, Yingqian Min, Beichen Zhang, Junjie Zhang, Zican Dong, et~al.
\newblock A survey of large language models.
\newblock \emph{arXiv preprint arXiv:2303.18223}, 2023.

\bibitem[Zheng et~al.(2025)Zheng, Zhou, Feng, Wang, and Lu]{zheng2025unicode}
Sipeng Zheng, Bohan Zhou, Yicheng Feng, Ye Wang, and Zongqing Lu.
\newblock Unicode: Learning a unified codebook for multimodal large language models.
\newblock In \emph{European Conference on Computer Vision}, pages 426--443. Springer, 2025.

\bibitem[Zhu et~al.(2024)Zhu, Ning, Jin, Lin, Huang, Song, Zhang, Tang, Pan, Zhou, et~al.]{zhu2024llmbind}
Bin Zhu, Munan Ning, Peng Jin, Bin Lin, Jinfa Huang, Qi Song, Junwu Zhang, Zhenyu Tang, Mingjun Pan, Xing Zhou, et~al.
\newblock Llmbind: A unified modality-task integration framework.
\newblock \emph{arXiv preprint arXiv:2402.14891}, 2024.

\bibitem[Zhu et~al.(2023)Zhu, Chen, Shen, Li, and Elhoseiny]{zhu2023minigpt}
Deyao Zhu, Jun Chen, Xiaoqian Shen, Xiang Li, and Mohamed Elhoseiny.
\newblock Minigpt-4: Enhancing vision-language understanding with advanced large language models.
\newblock \emph{arXiv preprint arXiv:2304.10592}, 2023.

\bibitem[Zhu et~al.(2025)Zhu, Zhang, Chen, Shi, Li, and Wu]{zhu2025connector}
Xun Zhu, Zheng Zhang, Xi Chen, Yiming Shi, Miao Li, and Ji Wu.
\newblock Connector-s: A survey of connectors in multi-modal large language models.
\newblock \emph{arXiv preprint arXiv:2502.11453}, 2025.

\end{thebibliography}
}

\end{document}